\documentclass[journal]{IEEEtran}
\usepackage{wrapfig}
\usepackage{xcolor}
\usepackage{amsthm}
\usepackage{graphicx}
\usepackage{multirow}
\usepackage{adjustbox}
\usepackage{bm}
\usepackage{bbold}

\usepackage{mathtools}

\usepackage{algorithm}
\usepackage{algorithmic}

\usepackage{float}
\usepackage{subfig}
\usepackage{wrapfig}
\usepackage{hyperref}
\usepackage{url}
\usepackage{graphicx} 

\usepackage{xurl}
\ifCLASSINFOpdf
\else
\fi

\begin{document}
%
\title{Cross-Domain Distribution Alignment for Segmentation of   Private Unannotated 3D   Medical Images}
%
%
%

\author{Ruitong Sun  and Mohammad Rostami, Member, IEEE
\thanks{S. Sun and M. Rostami are with the Department
of Computer Science, University of Southern California. }
\thanks{Manuscript received October, 2024.}}

%
%

\markboth{Journal of \LaTeX\ Class Files,~Vol.~14, No.~8, August~2015}%
{Shell \MakeLowercase{\textit{et al.}}: Bare Demo of IEEEtran.cls for IEEE Journals}
%



\maketitle

\begin{abstract}

Manual annotation of 3D medical images for segmentation tasks is  tedious and time-consuming. Moreover, data privacy  limits the applicability of  crowd sourcing to perform data annotation in medical domains. As a result, training deep neural networks for medical image segmentation can be challenging.
We introduce a new source-free Unsupervised Domain Adaptation (UDA)  method to address this problem. Our idea is based on estimating  the internally learned distribution of a relevant source domain by a base model and then  generating pseudo-labels that are used for enhancing the model refinement through self-training.
We demonstrate that our approach leads to SOTA performance on a real-world 3D medical  dataset.

\end{abstract}


%
\IEEEpeerreviewmaketitle

\section{Introduction}
\label{sec:introduction}

Automatic  semantic segmentation of medical images has seen significant advancements due to the emergence of models such as DeepLab \cite{chen2016semanticimagesegmentationdeep} and Vision Transformers (ViT)~\cite{dosovitskiy2020image}. A most notable model is the Segment Anything Model \cite{kirillov2023segment, ke2024segment}, which exemplifies the potential of transformer-based approaches in medical imaging, addressing the challenges posed by data scarcity and variability.
Despite these advancements,  adoption of these models to segmentation of 3D medical images remains limited due to the need for extensive, expertly annotated 3D datasets \cite{gangwal2024current}.
Unsupervised Domain Adaptation (UDA) \cite{wilson2020survey,nananukul2024multi} offers a promising solution.
UDA is different from zero-shot learning ~\cite{rostami2022zero} as it is assumed that
 unannotated data is accessible in a target domain to relax the need for annotated data by knowledge transfer from source domains with annotated data. Model adaptation is crucial in UDA for managing domain shifts that  degrade the source-trained model performance in new target domains. Traditional UDA methods rely on either adversarial training \cite{jian2023unsupervised} or metric minimization for distribution alignment \cite{gabourie2019learning}. Most UDA methods   require access to the source domain data which may not be feasible in medical domains due to the data privacy regulations.


We introduces a novel source-free UDA (SFUDA) method for segmentation of 3D medical imaging. Our idea  is to extract and utilize the internal distribution of the source domain data for domain alignment after training a model on the source domain. We   extract internal distributions from a trained model by cropping numerous subparts of the 3D source data, inputting these segments into the model, and saving the resulting distributions. For adaptation to the target data, we use a Gaussian Mixture Model (GMM) to estimate the internal distributions. This approach allows us to generate the closest pseudo-samples in the target domain which would facilitate effective alignment and enhance the source-train model performance for superior semantic segmentation on the target dataset. Our method circumvents the need for direct access to original training data, making it particularly advantageous in medical settings where sharing data  is not feasible. Our approach is validated on the Multi-Modality Whole Heart Segmentation Dataset \cite{zhuang2018multivariate} 3D dataset, achieving robust performance that exceeds conventional UDA techniques, despite maintaining the source domain data privacy

\section{Related work}
\label{sec:relatedwork}


\textbf{Semantic Segmentation in Medical Imaging}
Semantic segmentation   significantly enhances the clarity and interpretability of medical images \cite{strudel2021segmenter}. Traditionally, solving this task requires manual annotations by radiologists, which is labor-intensive and subject to variability.  The development of deep learning models, particularly convolutional neural networks (CNNs) led to a significant performance boost in this domain. Particularly, U-Nets \cite{ronneberger2015u}, an extension of CNNs, have become the common medical image segmentation models. Their encoder-decoder structure and skip connections help preserve spatial information. Vision Transformer (ViT) models \cite{kirillov2023segment} have further revolutionized this field by leveraging the power of transformers to handle diverse segmentation tasks with minimal training data. Despite these advancements, applying these models to 3D imaging remains challenging due to the need for extensive, expertly annotated 3D datasets.

\textbf{Unsupervised Domain Adaptation}
UDA  offers a promising solution to reduce the reliance on annotated data by adapting models trained on source domains with annotated data. This adaptation is crucial for mitigating domain shifts that   degrade model performance in new domains. Two mainstream techniques for domain adaptation are adversarial training and feature alignment using metric minimization.   
Both techniques  often require access to the source domain data which limits their adoption when the source domain data is  private. To address this challenge, Source-Free Unsupervised Domain Adaptation (SFUDA) method rely exclusively on   updating a  model, trained on the source data.   Recent approaches in SFUDA can be broadly categorized into two main streams. One stream focuses on self-training methods \cite{litrico2023guiding}, where pseudo-labels generated by the source model are used to iteratively adapt the model for the target domain through refining the model’s predictions. The other stream emphasizes the use of domain-specific normalization techniques \cite{kobler2022spd}, which adjust the model’s internal representations to better align with the target domain statistics. Both approaches aim to maintain high model accuracy while addressing the privacy and feasibility issues inherent in traditional domain adaptation methods.


\section{Problem Formulation}
\label{sec:problemformulation}

Our objective is to adapt a segmentation model,   trained using supervised learning on 3D annotated images of a source domain with distribution $\mathcal{S}$, to effectively generalize on a target domain with unannotated datat and a different data distribution $\mathcal{T}$, where $\mathcal{S} \neq \mathcal{T}$. The source domain dataset $\mathcal{D}^S = \{ (\bm{x}^s_1, \bm{y}^s_1), \ldots, (\bm{x}^s_n, \bm{y}^s_n) \}$ includes pairs of 3D images and their corresponding annotated masks. Here, $\bm{x}^s_i \in \mathbb{R}^{W \times H \times D}$ and $\bm{y}^s_i \in \{0,1\}^C$, where $W$, $H$, and $D$ denote the width, height, and depth of the images, respectively. The target domain dataset $\mathcal{D}^T$ comprises of only unannotated 3D medical images $\{ \bm{x}^t_1, \ldots, \bm{x}^t_m \}$ with the same sizes and semantic classes.
Annotations in $\mathcal{D}^S$ are crafted by clinical professionals, identifying specific anatomical or pathological features relevant to $C$ semantic classifications critical for clinical diagnostics. Externally, we know the same categorical semantic  classes
are helpful in the target domain.

We deploy a segmentation model, $M_\theta : \mathbb{R}^{W \times H \times D} \rightarrow \{0,1\}^C$, parameterized by $\theta$. This model is designed to map volumetric data into classification outputs for the $C$ predefined categories, prioritized according to their clinical significance. Notably, the SAM model, as delineated by Kirillov et al. \cite{kirillov2023segment}, illustrates how input data is converted into classifications, leveraging a transformer-based architecture that effectively handles diverse domain data.
To address model adaptation,
we consider that the model $\bm{M}_\theta$ comprises of three subnetworks—encoder $\bm{E}_\alpha$, decoder $\bm{D}_\beta$, and semantic segmentation classifier $\bm{S}_\phi$, with $\theta = \{\alpha, \beta, \phi\}$—and undergoes initial training on the source domain to enhance its generalization capabilities on the classless of interest. The encoder $\bm{E}_\alpha$ translates 3D medical images into latent space vectors, the decoder $\bm{D}_\beta$ refines these vectors and restores them to their original dimensions, and the classifier $\bm{S}_\phi$ translates these vectors into probabilistic segmentation maps with the same size of the input 3D images.

During initial training on the source domain, we utilize the cross-entropy (CE) loss for supervised learning:
\begin{equation}
\text{CE}(\bm{y}, \hat{\bm{y}}) = -\sum_{c=1}^C y_c \log(\bm{M}(\bm{x}^s)_c),
\end{equation}
where $\bm{y}$ is the one-hot encoded true label, and $\bm{M}(\bm{x}^s)$ represents the predicted probability for each class. 

After training on the source domain, the model needs to be adapted to generalize on the target domain because it exhibits variations in input data appearance due to domain-specific differences, e.g., MR vs CT images. Our goal is to enhance the model's performance on the target data without access to target labels during adaptation to preserve privacy. The solution for model adaptation involves aligning the source and target data distributions within the decoder's output space. The encoder $\bm{E}_\alpha$ and decoder $\bm{D}_\beta$, both with learnable parameters, transform the input data into a latent space $\mathbb{R}^{C}$. A distribution alignment function is then applied to minimize discrepancies, ensuring that the latent distributions from both domains become sufficiently similar to facilitate effective segmentation on the target domain, i.e.,   $ \phi$ generalizes on the target domain.

A common method for adapting to the target dataset involves using Generative Adversarial Networks (GANs) to morph source data to resemble target data, and vice versa, thereby learning features invariant across both domains. However, given the privacy constraints and the absence of annotated data in the target domain, our adaptation strategy must adhere to Source-Free Unsupervised Domain Adaptation (SFUDA) principles. The source-free setting restricts us to using either the source-trained model alone or transforming the source data into a non-reversible format, thereby ensuring that no direct access to the original source data is required during adaptation.

\section{Proposed 3D Medical Source-Free Model Adaptation Algorithm}

\begin{figure}[h!]
    \centering
    \includegraphics[width=\columnwidth]{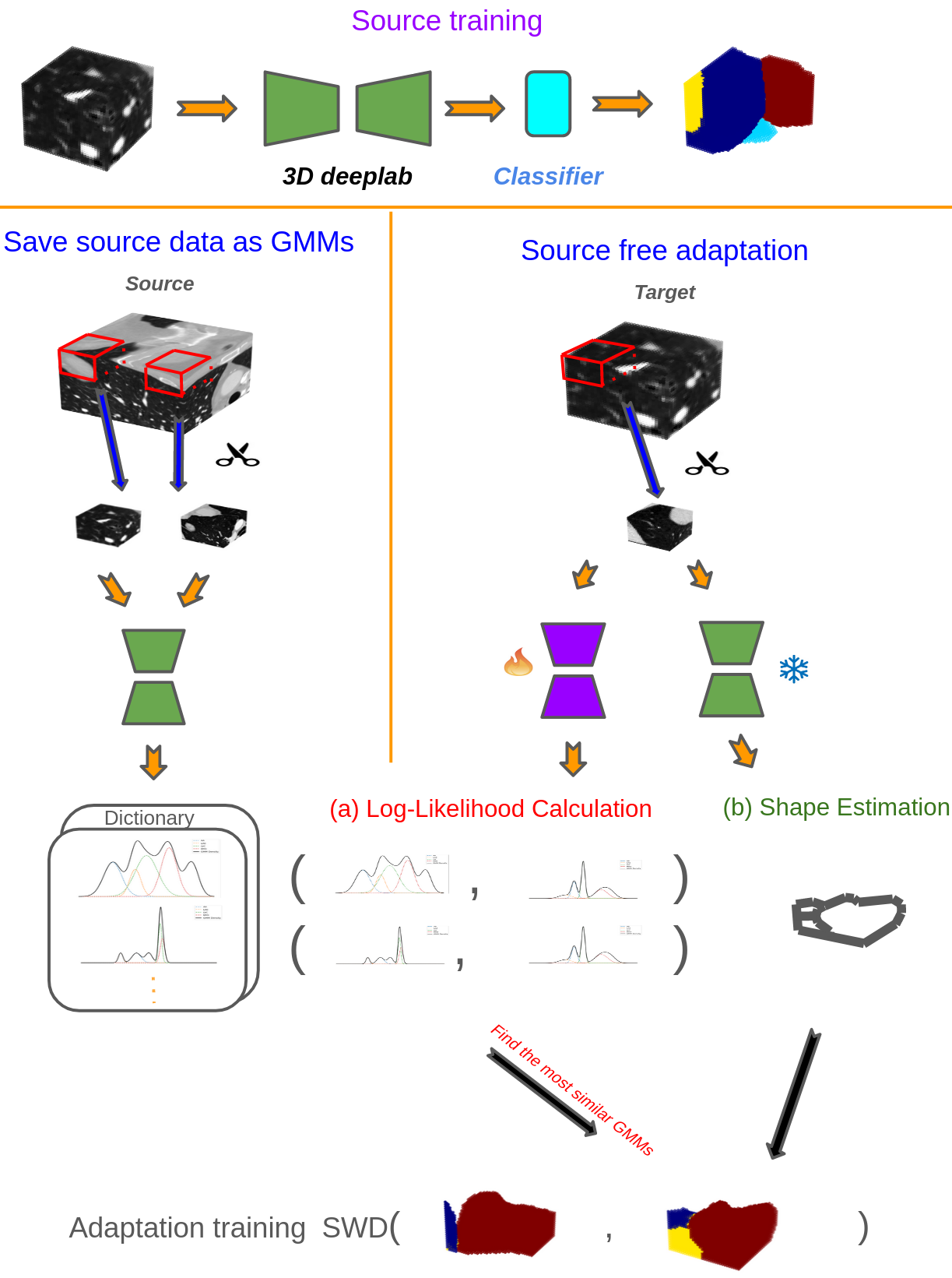}
    \caption{The proposed Model Adaptation  algorithm for 3D medical image segmentation. (top) the workflow begins with pre-training the segmentation model in a supervised setting using the source domain annotated samples. (bottom, left) to preserve privacy, the learned source distribution at the output-space of the encoder is estimated using a GMM. The GMM represents the internal feature distributions and is shared for model adaptation. (bottom, right)   the target domain images undergo   patch extraction and feature encoding. The purple encoder-decoder pair indicates a model initialized with parameters from the pre-trained source model. This model is adapted in two steps: (i) log-likelihood calculation compares the target domain GMM against the source domain GMM to find the most similar feature distributions; (ii) shape estimation involves feeding the target domain data into the frozen source-trained model to determine pixel-to-label assignments based on the source-trained parameters. These assignments provide an estimated shape for the target domain data, allowing for estimating the GMM latent features to be positioned close to these estimated shapes. The Sliced Wasserstein Distance   is then minimized as the loss function to align the target domain feature distribution with the the source domain distribution through the estimated GMM, enhancing segmentation quality.  }
    \label{fig:Architecture}
\end{figure}


Our solution involves aligning the source and the target domain distributions at the output space of the encoder indirectly to maintain domain privacy.
To this end,
we use the source-trained model to extract and estimate the internal data feature distribution of the source model using a Gaussian Mixture Models (GMM). Specifically, each source image $\bm{x}^s_i$ is cropped into patches of size $P_H \times P_W \times P_D$, and these patches are used to compute the latent representations $\bm{z}^s_{i,k}$, where $i$ indicates the sample index and $k$ indicates the cropping starting position (e.g., [0,0,0] for the initial point). These latent representations are then used to estimate GMM components $\bm{g}^s_{i,k}$ using the EM algorithm. The estimated GMM components are saved as a set $\mathcal{G}$. 
For model adaptation on the target domain, the target images are first cropped into patches $\bm{x}^t_{j,k}$, where $j$ denotes the sample index and $k$ the cropping start position. These patches are then fed into the $\bm{E} \circ \bm{D}$ from  adaptation model $\bm{M}$  to compute their latent representations $\bm{z}^t_{j,k}$, while simultaneously estimating the boundary of the target domain using the frozen pre-trained source model $\bm{M}^{\bm{S}}$. Once both the latent representations and the boundaries are obtained, they are combined by aligning the latent representations within their corresponding boundary regions, with each category assigned its own sub-boundary.

In parallel, the EM algorithm is applied to the latent representations to estimate the GMM $\bm{g}^t_{j,k}$ for the target domain features. This target GMM is then compared to the stored source domain GMMs in $\mathcal{G}$, using the largest log-likelihood metric to identify the most similar GMM. The latent representations from this most similar GMM, $\bm{g}^s_{i,k}$, are extracted and used as the target domain latent distribution to compute  the Sliced Wasserstein Distance (SWD) loss $\mathcal{L}_{\text{SWD}}(\bm{E} \circ \bm{D} (\bm{x}^t_{j,k}),  \bm{B}(\bm{g}^s_{i,k}))$, while the combined latent and boundary representation is used as the predicted latent.  SWD is a metric for distribution alignment \cite{lee2019sliced,standomain}.
By minimizing SWD, we ensure that distribution of the target domain in the latent aligns well with that of the source domain. This alignment facilitates the effective transfer of learned features from the source domain to the target domain, thereby  leading to acceptable segmentation performance on the target dataset. Algorithm~1 and Figure \ref{fig:Architecture}summarize our proposed approach.

\begin{algorithm}
\caption{Proposed Domain Adaptation Algorithm}
\label{alg:UDA}
\small
\begin{algorithmic}[1]
\REQUIRE Datasets: $\mathcal{D}^S = \{ (\bm{x}^s_i, \bm{y}^s_i) \}_{i=1}^n$, $\mathcal{D}^T = \{ \bm{x}^t_j \}_{j=1}^m$, Pre-trained model $\bm{M} = \bm{E} \circ \bm{D} \circ \bm{S}$

\STATE \textbf{Source   Training}
\FOR{each epoch}
    \FOR{each batch $(\bm{x}^s, \bm{y}^s)$ in $\mathcal{D}^S$}
        \STATE Preprocessing   $(\bm{x}^s, \bm{y}^s)$:  rotation, flipping,   scaling,  and cropping
        \STATE Forward Pass: $\hat{\bm{y}}^s =\bm{S}(\bm{D}( \bm{E}(\bm{x}^s)))$
        \STATE Compute   $\mathcal{L}_{\text{CE}}(\bm{y}^s, \hat{\bm{y}}^s) = -\sum_{c=1}^C y_c \log(\hat{y}_c)$
        \STATE Backward Pass: update $\bm{M}$   to minimize $\mathcal{L}_{\text{CE}}$
    \ENDFOR
\ENDFOR
\STATE \textbf{Privacy Preservation for Source-Free  Adaptation}
\FOR{each source image $\bm{x}^s_i$ in $\mathcal{D}^S$}
    \FOR{each crop $c_k$ from $\bm{x}^s_i$}
        \STATE Compute latent representation $\bm{z}^s_{i,k} = \bm{E}(c_k)$
        \STATE Decode latent representation $\bm{z}^{\prime s_{i,k}} = \bm{D}(\bm{z}^s_{i,k})$
        \STATE Apply the EM algorithm to obtain GMMs $\bm{g}^s_{i,k}$ from $\bm{z}^s_{i,k}$, then save the GMMs $\bm{g}^s_{i,k}$ to $\mathcal{G}$
    \ENDFOR
\ENDFOR
\STATE \textbf{Model Adaptation for the Target Domain}
\FOR{each batch $\bm{x}^t$ in $\mathcal{D}^T$}
    \STATE Crop $\bm{x}^t$ into patches  $c_k$ 
    \STATE Estimate boundary $\bm{Bs}$ = $\bm{B}(\bm{x}^t)$ using the  source-trained model $\bm{M}$
    \STATE Compute latent representation $\bm{z}^t = \bm{E}(\bm{x}^t)$
    \STATE Decode latent representation $\bm{z}^{\prime t} = \bm{D}(\bm{z}^t)$
    \STATE Apply EM to $\bm{z}^t$ to obtain target GMM $\bm{g}^t$
    \STATE Find the most similar GMM $\bm{g}^s_{i,k}$ to $\bm{g}^t$ from $\mathcal{G}$
    \STATE Organize the most similar GMM $\bm{g}^s_{i,k}$ embeddings within the estimated boundary
    \STATE Compute SWD   loss $\mathcal{L}_{\text{SWD}}(\bm{z}^t, \bm{Bs}(\bm{g}^s_{i,k}))$
    \STATE Update model parameters to minimize $\mathcal{L}_{\text{SWD}}$
\ENDFOR
\end{algorithmic}
\end{algorithm}

\section{Experiment Validation}
\label{sec: Experiment VALIDATION}

We validate our algorithm using  real-world medical image. Our implementation code is available to reproduce our results: \url{https://github.com/rusu4943/3D-SFUDA/tree/main}

\subsection{Experimental Setup}

\textbf{Dataset:}
Although there are several medical datasets for validating UDA algorithms for semantic segmentation, most of these datasets offer only 2D images. We evaluate our algorithm using the Multi-Modality Whole Heart Segmentation Dataset (MMWHS) \cite{gao2023bayeseg, zhuang2018multivariate, wu2022minimizing} which to best of our knowledge is the only publicly available 3D dataset, suitable for  UDA experiments. The raw 3D dataset includes seven segmentation classes: LV blood cavity (LV), RV blood cavity (RV), LA blood cavity (LA), RA blood cavity (RA), myocardium of the LV (MYO), AO trunk (AO), and PA trunk (PA). For a fair comparison with other UDA algorithms, we include only the ascending aorta (AA), left ventricle blood cavity (LVC), left atrium blood cavity (LAC), and myocardium of the left ventricle (MYO). The remaining three attributes, not covered in the literature, are treated as background class.
The MMWHS dataset contains CT and MR images, with each modality having 20 3D volumes which makes it suitable for UDA. In the source-free UDA setting for CT to MR, we train a model on 3D CT volumes and make these volumes inaccessible when adapting the model to work on MR images, and vice versa for the source-free setting from MR to CT. Unlike other papers that use 2D slices from 3D volumes, we use 3D volumes as input to preserve depth information, making the model's decisions benefit from spatial information in all directions.

\textbf{Preprocessing}
The MMWHS dataset was preprocessed through a comprehensive pipeline designed to standardize and prepare the data for effective segmentation.  
First, the label values are reset to ensure consistency across the dataset. Specific labels were remapped to have predefined classes, and non-target attributes were set to zero, focusing the segmentation task on the relevant anatomical structures.
Next, the image and label volumes were cropped based on the minimum and maximum coordinates of the non-zero labels, effectively isolating the heart region. This step reduces the data size and enhances the model’s efficiency by concentrating on the region of interest.
Subsequently, the image and label volumes were resampled to a consistent voxel size using trilinear interpolation for images and nearest neighbor interpolation for labels. This resampling step ensures uniform resolution across all samples, which is crucial for maintaining consistency during the training phase.
Intensity normalization was then applied to each sample. The top 2\% of the intensity histogram was clipped to eliminate outliers, followed by normalization of the intensities by subtracting the mean and dividing by the standard deviation. This process standardized the data distribution.
During the training phase on the source domain, data augmentation techniques were employed to enhance the model's robustness. This included random cropping to a size of \(32 \times 32 \times 32\), elastic transformations, color jittering, random rotations, scaling, translations, and flipping. These augmentations, except random cropping to a size of \(32 \times 32 \times 32\), were not applied during the adaptation phase to ensure consistency and reliability of the model's performance in target domain.

\textbf{Network Architecture:}
To leverage the spatial context and depth information in 3D medical images, we adapt the DeepLabV3 architecture   to a 3D version using VGG16 as the backbone. This adaptation involved replacing the 2D convolutions with the 3D convolutions, allowing the network to capture volumetric features essential for understanding spatial relationships in 3D medical images in all directions. Additionally, the network depth was reduced by one layer to prevent overfitting and to manage computational complexity.
The Atrous Spatial Pyramid Pooling (ASPP) module was incorporated in the 3D version, enabling  capturing of the multi-scale context through dilated convolutions with different rates. This feature improves the model's ability to understand objects at multiple scales, which is particularly beneficial for medical image segmentation. A global pooling layer was added to gather global context, which was then concatenated with the outputs from the ASPP module, enhancing segmentation accuracy.
Compared to a 3D U-Net, our 3D DeepLabV3 architecture provides a more robust handling of scale variations and captures more detailed contextual information due to its ASPP module. This ability makes DeepLabV3 particularly suitable for UDA, where the model needs to generalize well across different domains with varying characteristics. By exploiting the full volumetric information present in the medical images, 3D DeepLabV3 achieves more precise and reliable segmentation results, which is crucial for medical applications requiring an understanding of 3D anatomical structures.

\textbf{Evaluation:}
Following the literature, we use the Dice coefficient as our evaluation metric, defined as:

\[ \text{Dice} = \frac{2|A \cap B|}{|A| + |B|} \]

where \(A\) represents the set of predicted voxels, and \(B\) represents the set of ground truth voxels.   The Dice coefficient ranges from 0 to 1, where 1 indicates perfect overlap.

\textbf{Baselines for Comparison}

To the best of our knowledge, no prior method relies on 3D models for source-free UDA. Therefore, we compared our method against closely related approaches.
In our experiments, we have included classic 3D UDA algorithms: SIFA \cite{chen2020unsupervised} and SynSeg-Net \cite{huo2018synseg}. These methods have demonstrated effectiveness in medical image segmentation across different domains. We have also included the 2D source-free UDA method SFC \cite{stan2024unsupervised}. 
Furthermore, we extend our comparison to common UDA methods that are extended to work in the 3D setting through using our network architecture. Specifically, we include Cycle-Consistent Adversarial Domain Adaptation (cyCADA) \cite{hoffman2018cycada} and CycleGAN \cite{yang2018unpaired}.  For a fair comparison, we   compare against their 3D versions, including 3D-ADDA \cite{tzeng2017adversarial}, 3D-CycleGAN \cite{zhu2017unpaired}, 3D-cyCADA \cite{hoffman2018cycada}, and 3D-DICE \cite{zhang2019aet}. 
This diverse set of baselines allows us to thoroughly evaluate the performance of our proposed method and demonstrate that it is effective and leads to SOTA performance.

\subsection{Comparative Results}

Tables \ref{tab: Main result CT to MR} and \ref{tab: Main result MR to CT} provide the performance results for our algorithm along with the selected based lines. In each table, we have included the supervised learning performance on the target domain as an upperbound to evaluate the quality of UDA. We then have the 2D UDA methods. We expect improved performance over these methods because our method is relying on spatial relationships in all directions. However, we included these methods to demonstrate that using a 3D segmentation approach is beneficial compared to using 2D segmentation models.
Last, we have reported the performance of the 3D UDA methods. In each column, bold font denote the highest performance result. From the tables, we can observe that on average, the 3D UFA methods tend to outperform the 2D UDA methods. Particularly, comparing CyCADA and CycleGAN with their 3D counterparts, we observe that the same method performs better when 3D models are used for data processing. We can conclude that benefiting from 3D models in UDA is helpful. We also observe that our proposed approach outperforms the baselines for both tasks. It is important to note that our approach offers a source-free solution which is more challenging than the setting handled by most baselines. These results demonstrate that our approach is competitive and can reach  SOTA perofrmnace while maintaining the data privacy.


\begin{table}[h!]
\small
\centering
\begin{tabular}{|l|c|c|c|c|c|c|}
\hline
Method      & Dim   & \multicolumn{5}{c|}{Dice} \\ \cline{3-7} 
            &       & AA   & LAC  & LVC  & MYO  & Avg. \\ \hline
Supervised  & 3D    & 94.9 & 91.7 & 94.6 & 86.8 & 92.0 \\ \hline

SynSeg-Net  & 2D    & 71.6 & 69.0 & 51.6 & 40.8 & 58.2 \\ 
AdaOutput   & 2D    & 65.2 & 76.6 & 54.4 & 43.3 & 59.9 \\ 

CycleGAN    & 2D    & 73.8 & 75.7 & 52.3 & 28.7 & 57.6 \\ 
CyCADA      & 2D    & 72.9 & 77.0 & 62.4 & 45.3 & 64.4 \\ 

SIFA        & 2D    & 81.3 & 79.5 & 73.8 & 61.6 & 74.1 \\ 
\hline

3D-ADDA     & 3D    & 68.4 & 66.0 & 75.9 & 63.2 & 68.3 \\ 
3D-CycleGAN & 3D    & 60.9 & 70.8 & 73.6 & 63.1 & 67.1 \\ 
3D-CyCADA   & 3D    & 72.0 & 76.5 & 74.0 & 61.7 & 71.1 \\ 
3D-DISE     & 3D    & 81.2 & 87.0 & 83.6 & \textbf{68.4} & 80.3 \\ \hline

3D-SFUDA    & 3D    & \textbf{91.7} & \textbf{87.4} & \textbf{84.2} & 62.3 & \textbf{81.4} \\ \hline
\end{tabular}
\caption{Quantitative comparison for the UDA methods on the MR to CT adaptation task.}
\label{tab: Main result MR to CT}
\end{table}

\begin{table}[h!]
\small
\centering
\begin{tabular}{|l|c|c|c|c|c|c|}
\hline
Method      & Dim   & \multicolumn{5}{c|}{Dice} \\ \cline{3-7} 
            &       & AA   & LAC  & LVC  & MYO  & Avg. \\ \hline
Supervised  & 3D    & 77.0 & 86.5 & 92.5 & 75.8 & 83.0 \\ \hline

SynSeg-Net  & 2D    & 41.3 & 57.5 & 63.6 & 36.5 & 49.7 \\
AdaOutput   & 2D    & 60.8 & 39.8 & 71.5 & 35.5 & 51.9 \\ 

CycleGAN    & 2D    & 64.3 & 30.7 & 65.0 & 43.0 & 50.7 \\ 
CyCADA      & 2D    & 60.5 & 44.0 & 77.6 & 47.9 & 57.5 \\ 

SIFA        & 2D    & 65.3 & 62.3 & 78.9 & 47.3 & 63.4 \\ \hline

3D-ADDA     & 3D    & 33.2 & 44.6 & 74.9 & 53.9 & 51.7 \\ 
3D-CycleGAN & 3D    & 39.9 & 57.8 & 67.1 & 42.3 & 51.8 \\ 
3D-CyCADA   & 3D    & 49.1 & 66.2 & 76.8 & \textbf{55.9} & 62.0 \\ \hline

3D-SFUDA    & 3D    & \textbf{67.3} & 61.2 & 75.4 & 52.0 & \textbf{64.0} \\ \hline
\end{tabular}
\caption{Quantitative comparison for the UDA methods on the CT to MR  adaptation task.}
\label{tab: Main result CT to MR}
\end{table}

\begin{figure}[h]
    \centering
    \includegraphics[width=.9\columnwidth]{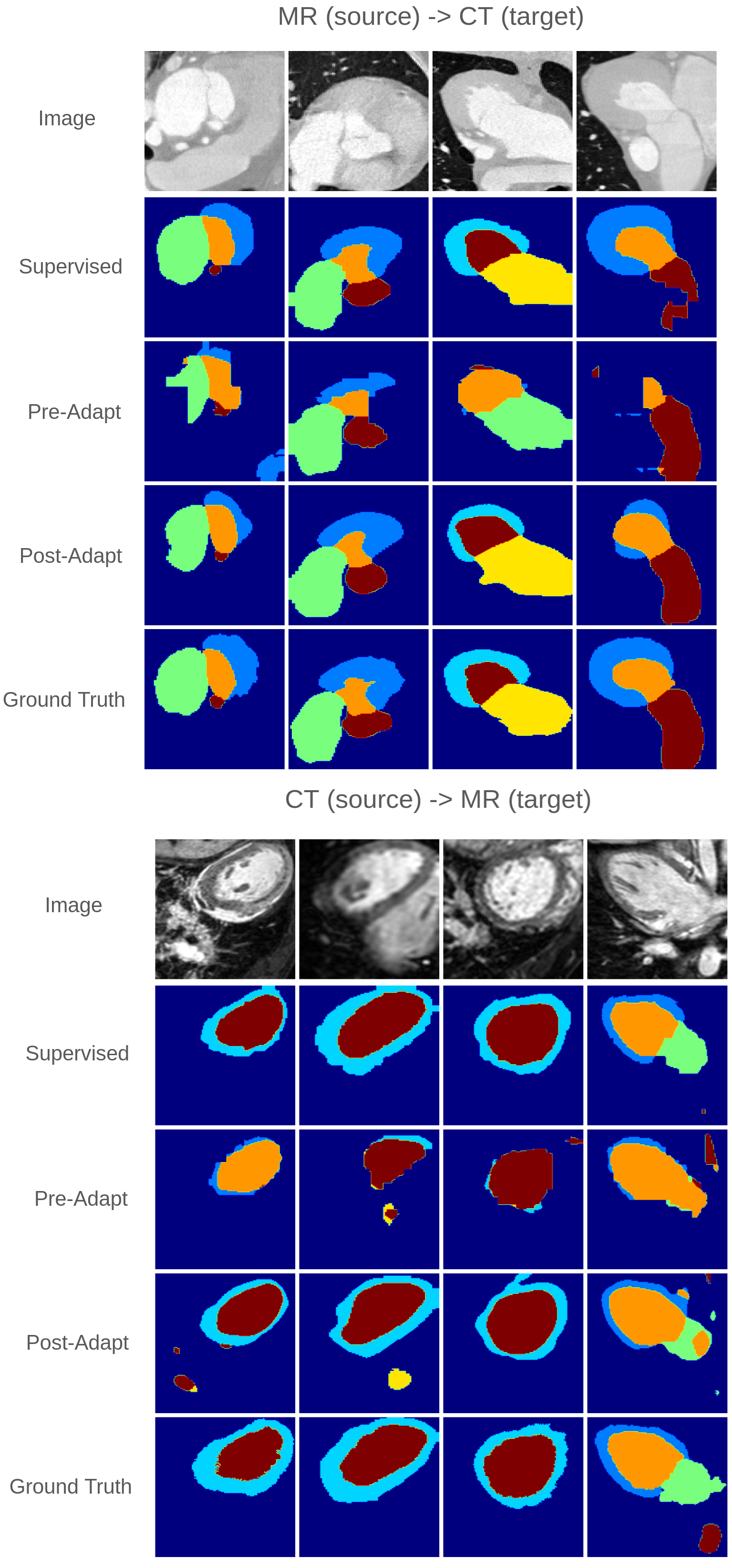}
    \caption{Qualitative performance: examples of the segmented frames for MMWHS dataset. From top to bottom in each case: input images, supervised learning predictions, source-trained model predictions, predictions based on our method, and ground truth provided by radiologists.}
    \label{fig:Main_results}
\end{figure}

To offer a more intuitive comparison, Figure \ref{fig:Main_results} provides an example of segmented images for each of the two tasks of the MMWHS dataset. In each figure, we can inspect visually the effect of using our method on the quality of semantic segmentation. Comparing the third and the fourth row in each case, we observe in both cases that using UDA can improve the performance of the source-model significantly to reach a performance similar to supervised learning (fifth row). We conclude that using UDA for semantic segmentation has the potential to relax the need for manual data annotation.

\subsection{Analytic Experiments}

Our proposed method comprises several key components, each critical for achieving optimal performance. One essential component is the search for the optimal GMM, which is used to compute the SWD loss for domain alignment.   Another vital component is the shape and count estimation of each label performed by the source model. Without any one of these elements, our method cannot achieve the best performance in the source-free UDA setting. 
We study the effect of these components on the final UDA performance.

\subsubsection{Searching for the optimal GMM}
\begin{figure}[h]
    \centering
    \includegraphics[width=\columnwidth]{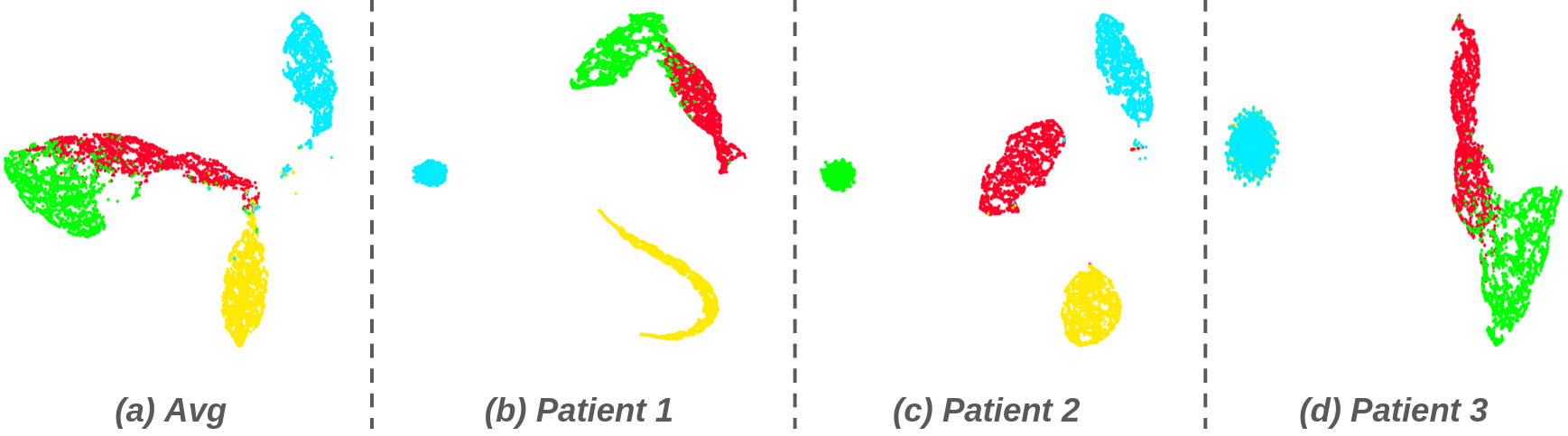}
    \caption{UMAP visualizations for  GMM samples: (a) the average for all patients' GMMs; (b) a GMM from one random crop of patient 1; (c) a GMMs from one random crop of patient 2; and (d) a GMMs from one random crop of patient 3. The colors represent different labels: red for MYO, yellow for LAC, green for LVC, and blue for AA. Cropping during adaptation is a benefit for adapted training because the target data crop sometimes contains only part of the labels.}
    \label{fig: Umap_searching_GMMs}
\end{figure}

To study the GMM estimation, we use UMAP \cite{mcinnes2020umapuniformmanifoldapproximation} for visualization tool. We aim to observe the differences between random crops from data that belong to different patients. As shown in Figure~\ref{fig: Umap_searching_GMMs}, subfigure ~\ref{fig: Umap_searching_GMMs} (a) illustrates the average GMM across all patients, displaying an almost equal distribution of pixel points for each color. In contrast, subfigures ~\ref{fig: Umap_searching_GMMs} (b), (c), and (d) represent random crops from individual patients, with each crop being $32 \times 32 \times 32$ in size (note that the data size is typically more than 100 in width and length and 50 in depth for each patient). Therefore, a crop may only contain parts of some labels.

In subfigure ~\ref{fig: Umap_searching_GMMs}  (b), the blue color representing the AA label has a smaller pixel count. In subfigure ~\ref{fig: Umap_searching_GMMs}  (c), the green color representing the LVC label in patient 2 exists in only a minimal number of pixels. Patient 3, shown in subfigure ~\ref{fig: Umap_searching_GMMs} (d), lacks any yellow pixels, indicating the absence of the LAC label due to cropping. In the best case for the adaptation training stage, the labels are ideally positioned in the same location in a patient with the same organ coverage. However, because the dataset lacks patients who provide both CT and MR images, there is no opportunity to find identical crops differing only in CT and MR. Therefore, we aim for the target domain by minimizing the SWD loss to have a similar count and shape for each label, which would help the model better understand the differences between CT and MR images.

\begin{table}[h!]
\small
\centering
\begin{tabular}{|l|c|c|c|c|c|}
\hline
\#GMM & \multicolumn{5}{c|}{Dice} \\ \cline{2-6} 
                       & AA   & LAC  & LVC  & MYO  & Avg. \\ \hline
Supervised             & 94.9 & 91.7 & 94.6 &  86.8 &  92.0 \\  \hline
  1           & 89.6 & 83.7 & 84.2 &  56.6 &  78.5 \\ 
  1298        & 90.6 & 84.6 & 85.3 &  57.6 &  79.5 \\ 
 27779       & 90.9 & 85.7 & 86.1 &  59.8 &  80.6 \\ \hline
\end{tabular}
\caption{Dice scores for MR to CT adaptation using varying numbers of GMM components.  Note that increasing the number of GMMs to 1,298 and 27,779 corresponds to decreasing the stride size to obtain more possible crops. The evaluation stride is set to (8,8,8) for these experiments.}
\label{tab: MR to CT; GMMs effect}
\end{table}

\begin{table}[h!]
\small
\centering
\begin{tabular}{|l|c|c|c|c|c|}
\hline
\#GMM & \multicolumn{5}{c|}{Dice} \\ \cline{2-6} 
                       & AA   & LAC  & LVC  & MYO  & Avg. \\ \hline
Supervised             & 77.0 & 86.5 & 92.5 & 75.8 & 83.0 \\ \hline
  1           & 50.1 & 59.4 & 73.0 & 64.4 & 61.7  \\ 
 1298        & 49.9 & 62.1 & 72.6 & 64.4 & 62.2 \\
  27779       & 50.6 & 59.5 & 74.0 & 64.2 & 62.1  \\ \hline
\end{tabular}
\caption{Dice scores for CT to MR adaptation with varying numbers of GMM components.   Note that increasing the number of GMMs to 1,298 and 27,779 corresponds to decreasing the stride size to obtain more possible crops. The evaluation stride is set to (8,8,8).}
\label{tab:  CT to MR; GMMs effect}
\end{table}

\textbf{MR to CT Adaptation:} In the MR to CT adaptation task (Table \ref{tab: MR to CT; GMMs effect}), the Dice scores demonstrate a clear upward trend as the number of GMM components increases from 1 to 27,779. This trend highlights the effectiveness of utilizing more GMM components to capture the variability present in the source domain data. Specifically, segmentation performance in key anatomical regions, such as the Ascending Aorta (AA) and Left Ventricle Cavity (LVC), shows significant improvement. For instance, the Dice score for the Myocardium (MYO) increases from 56.6 with a single GMM component to 59.8 with 27,779 GMM components.

\textbf{CT to MR Adaptation:} we observe that the CT to MR adaptation task (Table \ref{tab: CT to MR; GMMs effect}) does not exhibit the same level of improvement with an increased number of GMM components, as observed in the MR to CT direction. The Dice scores show only a modest improvement across different regions for classes  AA, LAC, LVC, and MYO, with slight fluctuations but no definitive upward trend. This observation suggests that the CT to MR adaptation might be inherently more challenging or less sensitive to changes in the number of GMM components. This phenomenon could be due to intrinsic differences in how CT data encapsulates anatomical information compared to MR data.
Additionally, we note that when the number of GMM components is drastically increased   27779 by a 10-fold value, we expected a further improvement in performance. However, the increased computational load limits the practicality of using this possibility.
We conclude that there is a trade-off for the value that we need to select for the GMM components.

\subsubsection{Shape \& Count estimation}

In our proposed algorithm, a key component is the shape estimation of the source model distribution. This process involves feeding the target input $x_t$ into the model pre-trained on the source domain, which provides an estimated region and a count for each label present in the target domain. Once these estimated regions and counts are obtained, we utilize the latent representations from the GMM that specifically represents the source domain. These GMM components encapsulate the latent space which contains information about semantic categories from the source domain. By using the count information, we insert the relevant latent spaces from the GMM into the estimated target label regions. This targeted approach ensures that the segmentation model aligns closely with the inherent structure of the target domain. In Tables \ref{tab: CT to MR; shape estimation} and  \ref{tab: MR to CT; shape estimation}, we have provided the performance of models without shape estimation and count estimation  compared to the complete 3D-SFUDA pipeline. The evaluation stride is set to (8,8,8) with a single GMM used for estimation.
We observe that the impact of utilizing proper counts and shape  is notable. For the task of adapting from MR to CT, our proposed approach results in a 4\% increase in the Dice coefficient. When adapting from CT to MR, there is a 1.4\% increase in the Dice coefficient, which still is an improvement, and further demonstrate the efficacy of our method in enhancing segmentation performance across different imaging modalities.

\begin{table}[h!]
\small
\centering
\begin{tabular}{|l|c|c|c|c|c|}
\hline
Method               & \multicolumn{5}{c|}{Dice} \\ \cline{2-6} 
                     & AA   & LAC  & LVC  & MYO  & Avg. \\ \hline
w/o Shape + Count    & 50.6 & 59.3 & 68.2 & 64.3 & 60.6 \\
w/o Shape            & 50.5 & 59.2 & 68.7 & 64.4 & 60.7 \\
3D-SFUDA             & 49.9 & 62.1 & 72.6 & 64.4 & 62.2 \\ \hline
\end{tabular}
\caption{Ablative study of the effect of the shape and count estimation in the adaptation process for MR to CT task. }
\label{tab: CT to MR; shape estimation}
\end{table}

\begin{table}[h!]
\small
\centering
\begin{tabular}{|l|c|c|c|c|c|}
\hline
Method & \multicolumn{5}{c|}{Dice} \\ \cline{2-6} 
                              & AA   & LAC  & LVC  & MYO  & Avg. \\ \hline
w/o Shape + Count             & 90.5 & 82.7 & 83.0 & 49.9 & 76.5 \\
w/o Shape                     & 90.6 & 82.7 & 83.1 & 50.2 & 76.6 \\
3D-SFUDA                      & 91.0 & 85.7 & 86.1 & 59.8 & 80.6 \\ \hline
\end{tabular}
\caption{Ablative study of the effect of the shape and count estimation in the adaptation process for  MR to CT task.}
\label{tab: MR to CT; shape estimation}
\end{table}

\begin{figure}[h]
    \centering
    \includegraphics[width=\columnwidth]{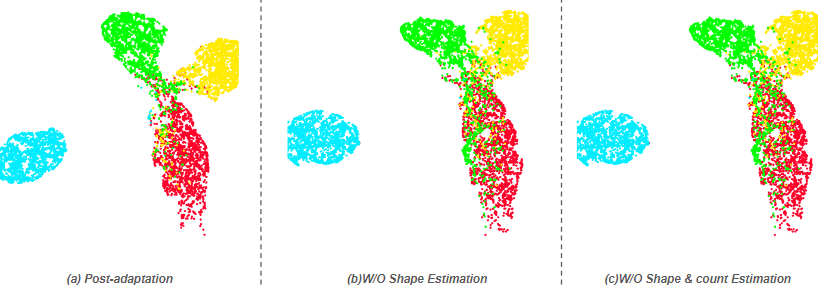}
    \caption{Comparison of internal latent distributions during post-adaptation: (a) the internal latent distribution after post-adaptation using both shape and count estimation; (b) the internal latent distribution after post-adaptation without shape estimation but using the estimated count of each label; (c) the internal latent distribution after post-adaptation without both shape estimation and estimated count of each label. The colors represent different labels: red for MYO, yellow for LAC, green for LVC, and blue for AA. It can be observed that cropping during adaptation enhances training by focusing on label subsets present in the target data crop.}
    \label{fig: shape estimation}
    \vspace{-6mm}
\end{figure}

In Figure \ref{fig: shape estimation}, we have visualized the internal distribution of the data during the adaptation process using UMAP. We can observe a  trend similar to the previous tables, further demonstrating the effect of shape and count estimation on the quality of estimating the source domain.

\subsubsection{Effect of UDA on the internal data distribution}
\begin{figure}[h]
    \centering
    \includegraphics[width=\columnwidth]{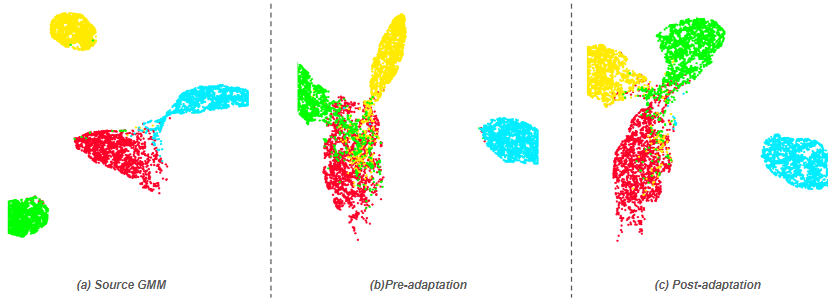}
    \caption{Comparison of internal latent distributions during different adaptation phases: (a) the internal latent distribution with input source data before adaptation; (b) the internal latent distribution with input target data before adaptation; (c) the internal latent distribution with input target data after adaptation. The colors represent different labels: red for MYO, yellow for LAC, green for LVC, and blue for AA. }
    \label{fig: Pre_and_Post internal distribution}
\end{figure}

We provide an intuitive visualization to demonstrate the effectiveness of our approach by studying the internal data distribution. To this end, we visualize the data distribution in the shared embedding space. As shown in Figure \ref{fig: Pre_and_Post internal distribution}, subfigure \ref{fig: Pre_and_Post internal distribution} (a) depicts the internal latent distribution of the source dataset when fed into the pre-adaptation model. Here, we observe minimal overlap between the different classes, indicating well-separated clusters, which serves as a benchmark for the best internal distribution structure post-adaptation. Subfigure \ref{fig: Pre_and_Post internal distribution} (b) presents the internal latent distribution when the target data is input into the pre-adaptation model, where significant overlap between clusters of different classes is evident. This overlap suggests that the model struggles to distinguish between different labels when applied to the target domain data, highlighting that the domain shift problem exists. Subfigure \ref{fig: Pre_and_Post internal distribution} (c) shows the internal latent distribution after adaptation when the target data is fed into the post-adaptation model. Compared to Subfigure \ref{fig: Pre_and_Post internal distribution} (b), the overlap between classes is notably reduced, demonstrating the effectiveness of our approach in aligning the source and target distributions. This reduction in overlap signifies improved class separation and indicates successful adaptation to the target domain, achieving the desired internal distribution similar to the source as shown in Subfigure \ref{fig: Pre_and_Post internal distribution} (a).

\subsubsection{ Stride value during evaluation }

 Tables \ref{tab:  MR to CT; stride effect}   and \ref{tab:  CT to MR; stride effect} illustrate the impact of different stride sizes on the Dice scores in domain adaptation using 3D-SFUDA. The numbers in parentheses represent the stride size for each experiment. As the stride decreases from \(16 \times 16 \times 16\) to \(2 \times 2 \times 2\), there is a consistent improvement in segmentation accuracy across both tasks. 
 These results highlight a clear trade-off in that smaller strides enhance features and improve segmentation accuracy but also increase the computational load significantly. This trade-off underscores the need for balance between computational efficiency and segmentation precision, particularly in clinical applications where processing speed is crucial.

\begin{table}[h!]
\small
\centering
\begin{tabular}{|l|c|c|c|c|c|}
\hline
Method & \multicolumn{5}{c|}{Dice} \\ \cline{2-6} 
                       & AA   & LAC  & LVC  & MYO  & Avg. \\ \hline
Supervised             & 94.9 & 91.7 & 94.6 & 86.8 & 92.0 \\  \hline
3D-SFUDA (16,16,16)    & 90.9 & 84.4 & 83.1 & 55.9 & 78.6  \\ \hline
3D-SFUDA (8,8,8)       & 91.0 & 85.7 & 86.1 & 59.8 & 80.6  \\ \hline
3D-SFUDA (4,4,4)       & 91.3 & 87.0 & 86.4 & 61.0 & 81.4  \\ \hline
3D-SFUDA (2,2,2)       & 91.7 & 87.4 & 84.2 & 62.3 & 81.4  \\ \hline
\end{tabular}
\caption{Impact of   stride size on performance of 3D-SFUDA for the MR to  CT task. Each stride, shown in parentheses, is evaluated using 1298 GMM components.}
\label{tab: MR to CT; stride effect}
\end{table}

\begin{table}[h!]
\small
\centering
\begin{tabular}{|l|c|c|c|c|c|}
\hline
Method & \multicolumn{5}{c|}{Dice} \\ \cline{2-6} 
                       & AA   & LAC  & LVC  & MYO  & Avg. \\ \hline
Supervised             & 77.0 & 86.5 & 92.5 & 75.8 & 83.0 \\ \hline
3D-SFUDA (16,16,16)    & 48.0 & 56.6 & 71.4 & 62.6 & 59.6 \\ \hline
3D-SFUDA (8,8,8)       & 49.9 & 62.1 & 72.6 & 64.4 & 62.2 \\ \hline
3D-SFUDA (4,4,4)       & 51.5 & 63.1 & 73.4 & 65.7 & 63.4 \\ \hline
3D-SFUDA (2,2,2)       & 67.3 & 61.2 & 75.4 & 52.0 & 64.0 \\ \hline
\end{tabular}
\caption{Impact of   stride size on performance of 3D-SFUDA for the   CT to MR task. Each stride, shown in parentheses, is evaluated using 1298 GMM components.}
\label{tab:  CT to MR; stride effect}
\end{table}

\begin{figure}[h]
    \centering
    \includegraphics[width=.7\columnwidth]{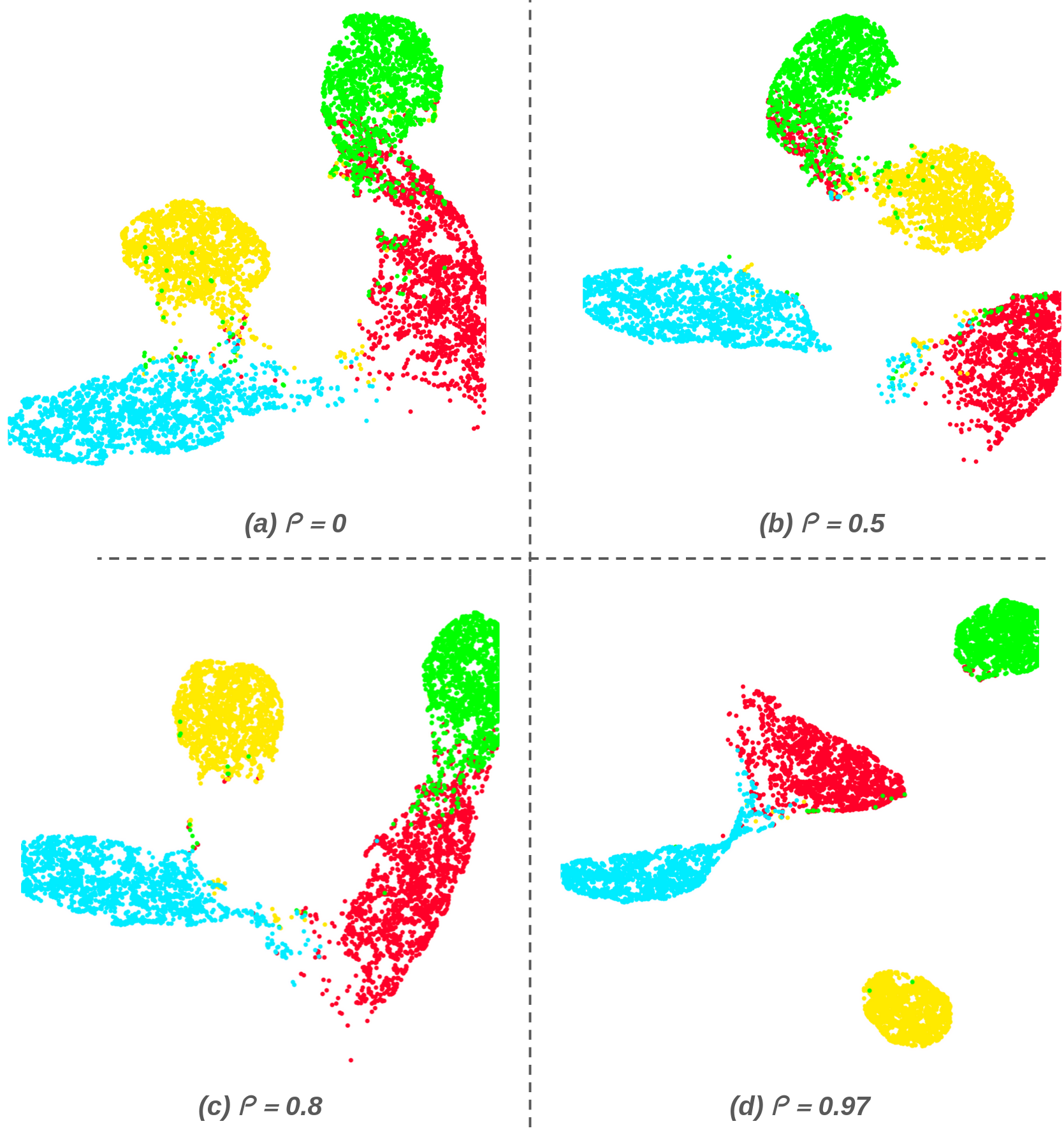}
    \caption{Impact of the $\rho$ parameter on the internal distribution. }
    \label{fig: rou_parameter}
\end{figure}

\subsubsection{The impact of the $\rho$ parameter} Figure \ref{fig: rou_parameter} demonstrates how different values of $\rho$ affect the internal representation of source data. As $\rho$ increases, the different labels become more distinguishable from each other, which is beneficial for the adaptation phase. A clearer separation of latent representation allows for more effective learning and adaptation, improving the overall performance of the model.

\section{Conclusion}

We demonstrated that our 3D-SFUDA method effectively addresses privacy constraints in 3D medical imaging. Our approach was validated on the MMWHS dataset, showing robust performance in both CT-to-MRI and MRI-to-CT adaption directions. Additionally, we conducted ablation experiments to evaluate the contribution of each component of our method, confirming their individual importance. Furthermore, our analysis highlighted the critical role of parameter selection in achieving successful outcomes. These findings underscore the effectiveness of our approach in adapting medical imaging models while respecting data privacy concerns.

\small

\bibliographystyle{IEEEtran}
\bibliography{ref}

\end{document}